\documentclass[preprint,12pt]{elsarticle}

\usepackage{amssymb}

\usepackage{pifont}
\newcommand{\cmark}{\ding{51}}%
\newcommand{\xmark}{\ding{55}}%
\usepackage{booktabs}
\usepackage{natbib}
\usepackage{rotating}
\usepackage{xcolor}

\journal{Knowledge-Based Systems}

\begin{document}

\begin{frontmatter}



\title{Capturing and Anticipating User Intents in Data Analytics via Knowledge Graphs}

\author[UPC]{Gerard Pons\corref{cor1}}
\ead{gerard.pons.recasens@upc.edu}
\author[UPC]{Besim Bilalli}
\ead{besim.bilalli@upc.edu}
\author[UPC]{Anna Queralt}
\ead{anna.queralt@upc.edu}

\affiliation[UPC]{organization={Universitat Politècnica de Catalunya},
            city={Barcelona},
            country={Spain}}

\cortext[cor1]{Corresponding author}

\begin{abstract}
In today's data-driven world, the ability to extract meaningful information from data is becoming essential for businesses, organizations and researchers alike. For that purpose, a wide range of tools and systems exist addressing data-related tasks, from data integration, preprocessing and modeling, to the interpretation and evaluation of the results. As data continues to grow in volume, variety, and complexity, there is an increasing need for advanced but user-friendly tools, such as intelligent discovery assistants (IDAs) or automated machine learning (AutoML) systems, that facilitate the user’s interaction with the data. This enables non-expert users, such as citizen data scientists, to leverage powerful data analytics techniques effectively.

The assistance offered by IDAs or AutoML tools should not be guided only by the analytical problem's data but should also be tailored to each individual user. To this end, this work explores the usage of Knowledge Graphs as a basic framework for capturing in a human-centered manner complex analytics workflows, by storing information not only about the workflow's components, datasets and algorithms but also about the users, their intents and their feedback, among others. The data stored in the generated Knowledge Graph can then be exploited to provide assistance (e.g., recommendations) to the users interacting with these systems. To accomplish this objective, two methods are explored in this work. Initially, the usage of query templates to extract relevant information from the Knowledge Graph is studied. However, upon identifying its main limitations, the usage of link prediction with knowledge graph embeddings is explored, which enhances flexibility and allows leveraging the entire structure and components of the graph. The experiments show that the proposed method is able to capture the graph's structure and to produce sensible suggestions. To demonstrate the feasibility of the approach, a prototype is presented.

\end{abstract}



\end{frontmatter}


\section{Introduction}
\label{sec:intro}
Machine Learning (ML) is a branch of Artificial Intelligence which focuses on building algorithms from data that can be used to perform different complex tasks (e.g., identify hidden patterns, make predictions, provide suggestions, etc.). This conversion of data into knowledge has numerous applications across multiple fields and industries, and its relevance is growing with the increasing availability and abundance of data. However, since transforming raw data into valuable information can be time-consuming and challenging, various efforts have been done to simplify some of the laborious steps of the ML workflow creation for experts, while enabling inexperienced users to construct valuable models without requiring significant programming or ML expertise. For instance, Intelligent Discovery Assistants (IDAs) are used to interactively assist users while creating Data Analytics (DA) workflows. To this end, they leverage expert derived rules, previously successful workflows, metadata about the input dataset or about the operators to propose workflows~\cite{idas}. Another approach is AutoML~\cite{automl}, which is the automation of various ML stages by looking for the best workflow (or workflow component) by optimizing the analytical problem over a search space of algorithms and parameters.

Using these tools, however, is not completely straightforward for non-technical users as various decisions still have to be made throughout the whole process (e.g., selecting the analytical purpose, identifying a suitable metric, etc.). These decisions can have a huge impact on the quality of the results, hence guiding users while generating ML workflows can improve the effectiveness of the analysis and make the whole process more efficient~\cite{GIOVANELLI2022101957}. However, generating these recommendations is not a trivial task, as they should depend on the user's characteristics (e.g., area of expertise, role in a company), user's past experiences, datasets used, etc. 

To this end, this work proposes a method to provide recommendations for the users' decisions by leveraging the information captured in a Knowledge Graph (KG). KGs are a form of knowledge representation that organize information in a structured and interconnected manner by representing a network of entities and the different relations between them. One of the standard ways of defining a KG is by expressing relationships as a set of three elements called a triple \textit{(h,r,t)}, consisting of the head entity (h), the relation (r), and the tail entity (t). This semantically rich representation preserves structural information while making it both machine-readable and easily understandable by humans.

\begin{figure}[htpb]
    \centering
    \includegraphics[width=0.9\textwidth]{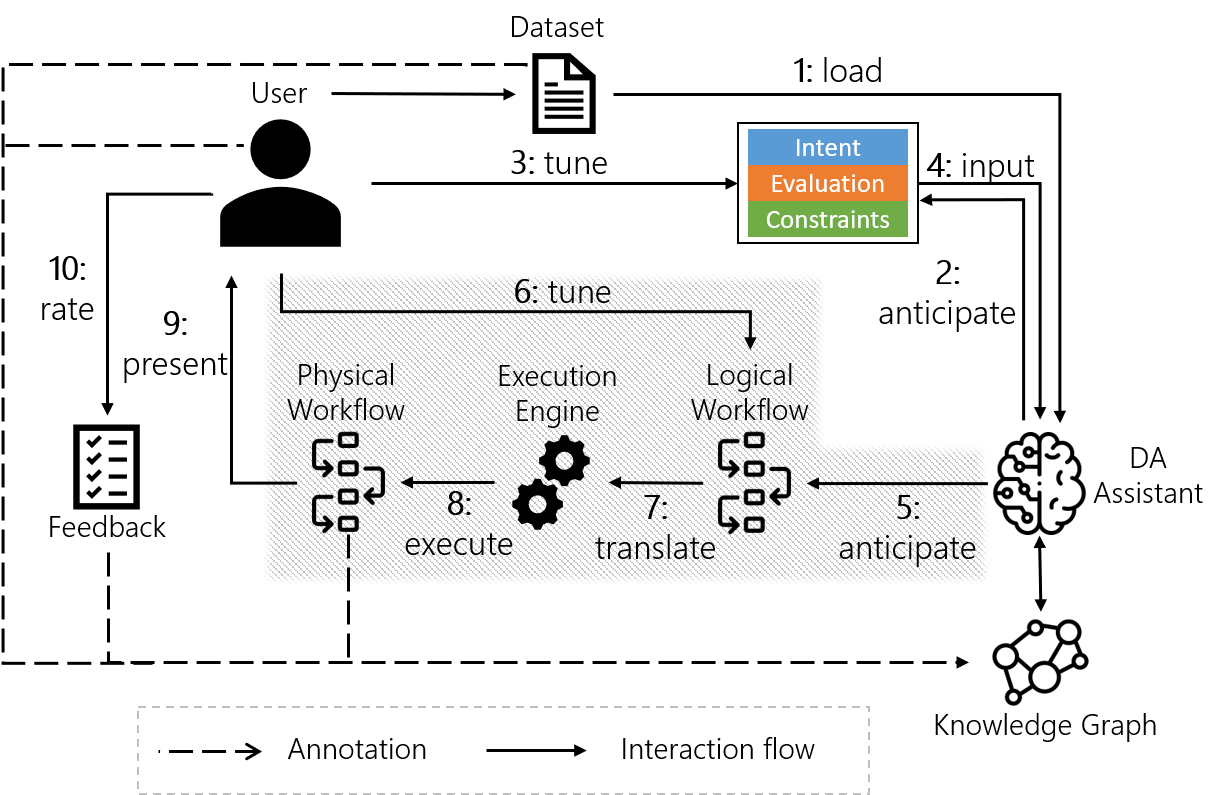}
  \caption{Overview of a DA Assistant interacting with a KG. In this work, the grayed steps have been automated via an AutoML tool.}
  \label{fig:general}
\end{figure}

Figure~\ref{fig:general} provides the context of how the anticipation system developed in this work can be used to assist the user in the initial phase of defining the input in different DA Assistants. For that, we assume that the assistant can interact with a KG, which contains annotations of the users' past interactions with the system. To this end, after uploading a dataset (arrow 1), the system can provide recommendations (arrow 2) for the users' input (e.g., intent, constraints, etc.) based on information about the dataset, the users and previous experiments stored in the KG. These are the recommendations that fall under the scope of this work and they can be reviewed and changed by the users (arrow 3) to finally define the input to the system (arrow 4). Next, considering the case of an IDA~\cite{dorian, IDAmeta}, the system would suggest logical workflows (arrow 5), which the users can tune (arrow 6) before their execution (arrows 7 and 8). In the case of AutoML systems~\cite{autosklearn,tpot,hyperopt}, steps 5-8 (highlighted with a gray box) are replaced by an automatic optimization of the process based on the specified input. In both cases, the users are presented with the results (arrow 9) and can provide their feedback (arrow 10).

In this work, we instantiate the interaction shown in Figure~\ref{fig:general} and design a KG for the DA domain. However, in addition to capturing the different components of analytical workflows, their characteristics and the relationships between them, the KG captures information about the users, their feedback over the workflows and the intent for which these workflows have been generated. Now, to transform the information stored in the KG into user-centered recommendations (see Figure~\ref{fig:general}, arrow 2), we first explore how the KG structure can be exploited by using predefined query templates. These templates allow the generation of simple but sensible recommendations, and are designed taking domain knowledge into account. However, this approach lacks flexibility and does not use the available information of the graph unless complex ad-hoc queries are posed for every request. To address this problem, we exploit the whole structure and components of the KG by creating embeddings for its different elements. With them, we build a recommender system for the user inputs based on link prediction, that enables leveraging all the information stored in the graph. To this end, the main contributions of this work are:
\begin{itemize}
    \item We develop a KG to capture and store the human-in-the-loop (i.e., users, intents, preferences, constraints) together with the analytical artifacts (e.g., datasets, algorithms, workflows) required for the automatic generation of analytical workflows.
    \item We propose how to represent DA intents with a hierarchical structure with different levels of abstraction, going from the most general intents (e.g., describe, suggest, predict, etc.) to concrete algorithms and their implementations.
    \item We explore the use of query templates to retrieve information from the KG to provide human-in-the-loop assistance when generating analytical workflows.
    \item We propose a more flexible method, based on graph embeddings, that leverages the KG's topology to anticipate and recommend the user input by using link prediction.
    \item We evaluate our approach to show its feasibility and present a prototype to demonstrate its potential use in practice. 
\end{itemize}

The remainder of this paper is organized as follows. In Section~\ref{sec:related_work}, we discuss the Related Work. In Section~\ref{sec:kg_design}, we detail the design of the KG and in Section~\ref{sec:population} we describe how it has been populated. In Section~\ref{sec:anicipation}, we discuss how the KG can be exploited to provide recommendations for the user input (i.e., intentions, constraints and preferences) by querying the KG and by using knowledge graph embeddings. For the latter, we provide an experimental evaluation in Section~\ref{sec:experiments}. Finally in Section~\ref{sec:prototype}, a prototype implementing both methods is presented, and in Section~\ref{sec:conclusion} we provide the conclusions and ideas for future work.
\section{Related Work}
\label{sec:related_work}

In the past decade, different works have developed KGs or ontologies\footnote{Ontologies are KGs that formally make a distinction between the schema and the actual data, providing means of inference and reasoning.} to capture ML workflows and experiments. Additionally, some works have also studied the usage of such ontologies to enhance automatic decision making in the different steps of the DA process. In the following sections, the most relevant works are introduced, and the differences with our approach are stated. 

\subsection{Capturing Data Analytics via Ontologies}
Various ontologies have been created to capture different parts of the DA process. In~\cite{dmop}, the DMOP ontology is designed to assist in making well-informed decisions at different stages of the DA process. It contains descriptions (e.g., assumptions, properties, optimization strategies, etc.) of DA tasks, workflows, algorithms, hypotheses and data. In~\cite{ontodm}, the OntoDM ontology is presented, defining a general DA space represented in a three-layered structure: the specification, implementation and application layer. In~\cite{bigowl} the BIGOWL ontology is presented to support knowledge management in Big Data analytics, by representing how different components are connected in ML workflows and how the generated data is transferred between them. Also, in~\cite{mlschema}, the ML-Schema ontology  is designed as a top-level ontology to represent ML experiments, with their algorithms and datasets. It is designed to align more fine-grained DA ontologies, such as OntoDM or DMOP. 

To the best of our knowledge, there does not exist an ontology which incorporates the users, their input constraints and preferences, and their feedback (see Table~\ref{tab:comparison}), which are necessary for a better understanding of the DA process and enable intentional analytics~\cite{intentional} in ML.

\begin{table*}[]
\caption{Comparison of relevant classes available in the studied ontologies.}
\label{tab:comparison}
\small
\begin{tabular}{cccccccc}
\toprule
 & Workflow & Data & Algorithm & Eval. & Constr. & Intent & User \\
 \midrule
OntoDM & \cmark & \cmark & \cmark & \cmark & \cmark & \xmark & \xmark \\
DMOP &  \cmark & \cmark & \cmark & \cmark & \xmark & \xmark & \xmark \\
ML-Schema & \cmark & \cmark & \cmark & \cmark & \xmark & \xmark & \xmark \\
BIGOWL & \cmark & \cmark & \cmark & \cmark & \xmark & \xmark & \xmark \\
\bottomrule
\end{tabular}
\end{table*}

\subsection{Anticipating Data Analytics via Ontologies}
In the context of DA Assistants, some of the enumerated ontologies have been used to assist the users in the generation of ML workflows. In~\cite{IDAmeta}, a set of hierarchical task networks (HTN) representing workflows are extended using the DMOP ontology, and frequent patterns are extracted using a tree mining algorithm. These frequent patterns are then used to represent each generated workflow. At the same time, the datasets used in each workflow are represented by some defined characteristics (e.g., number of instances, proportion of missing values, etc.). Finally, by computing similarities between datasets, workflows or between the workflow-dataset pair, different workflows can be generated when a new dataset is to be processed. Similarly, in~\cite{dmop}, DA experiments annotated with DMOP are used to create a probabilistic ranker, which returns an ordered list of workflows based on the expected performance. Also with a ranking approach,~\cite{ida} proposes adding general estimated qualities over an ontology of DA algorithms (such as speed, accuracy or memory) and then generates workflows ranked based on the selected criterion.  

The presented workflow automation systems focus on capturing the DA process, with emphasis on the different workflow's elements and their attributes, but they again do not take into account the end users' characteristics, their past analyses or their subjective evaluation (feedback) of the results. In our work, we incorporate such concepts into the KG, and we leverage them to enable the creation of workflows that are more relevant to the end user’s needs.
\section{Knowledge Graph Design}
\label{sec:kg_design}
The goal of this work is to design a KG capable of capturing ML workflows together with users' information, which then will be leveraged with the objective of facilitating the creation of analytical workflows. First, the new concepts that have been included in the KG are introduced, with special attention to the user intents. Then, we explain the domain and scope of the graph, the reused resources for its creation and finally the design of its classes and properties. The resulting KG will define the expressivity considered in the rest of the paper. 

\subsection{Preliminaries}
\label{sec:user_input}
Let us start with a brief explanation of the new user’s input classes that are being considered in this work, which are not present in the ontologies presented in Section~\ref{sec:related_work} (see Table~\ref{tab:comparison}), and that will be modeled in the KG.
\subsubsection{User Intents}
User intents in the context of data analysis state, at different levels of abstraction, the main goal of the analysis in which the user is interested without specifying the details of how it should be accomplished. Correctly identifying user intents is crucial for the generation of fit-for-purpose workflows. In a high level view, intents can be classified in five different groups~\cite{intentional}:
\begin{itemize}
    \item \textbf{Describe}: the aim is to provide a description of the instances or variables of the dataset. For example, clustering techniques or outlier detection methods could be used for this purpose.
    \item \textbf{Assess}: the focus is on providing assessment of the performance of different processes or on comparing them to a baseline. Therefore, techniques such as benchmarking or KPI assessment are usually applied for this purpose.
    \item \textbf{Explain}: the aim is to provide an explanation of the relations between instances and/or variables, usually by capturing the hidden relationships between them.  Methods related to statistical tests are the most common in this category.
    \item \textbf{Predict}: the goal is to forecast a future event or to predict the likelihood of a particular outcome. Classification and Regression algorithms can be found in this group.
    \item \textbf{Suggest}: the aim is to provide recommendations for future interactions. Therefore, Recommender Systems or Association Rule Learning are used to achieve this purpose.
\end{itemize}

\begin{figure}[htpb]
    \centering
    \includegraphics[width=1\textwidth]{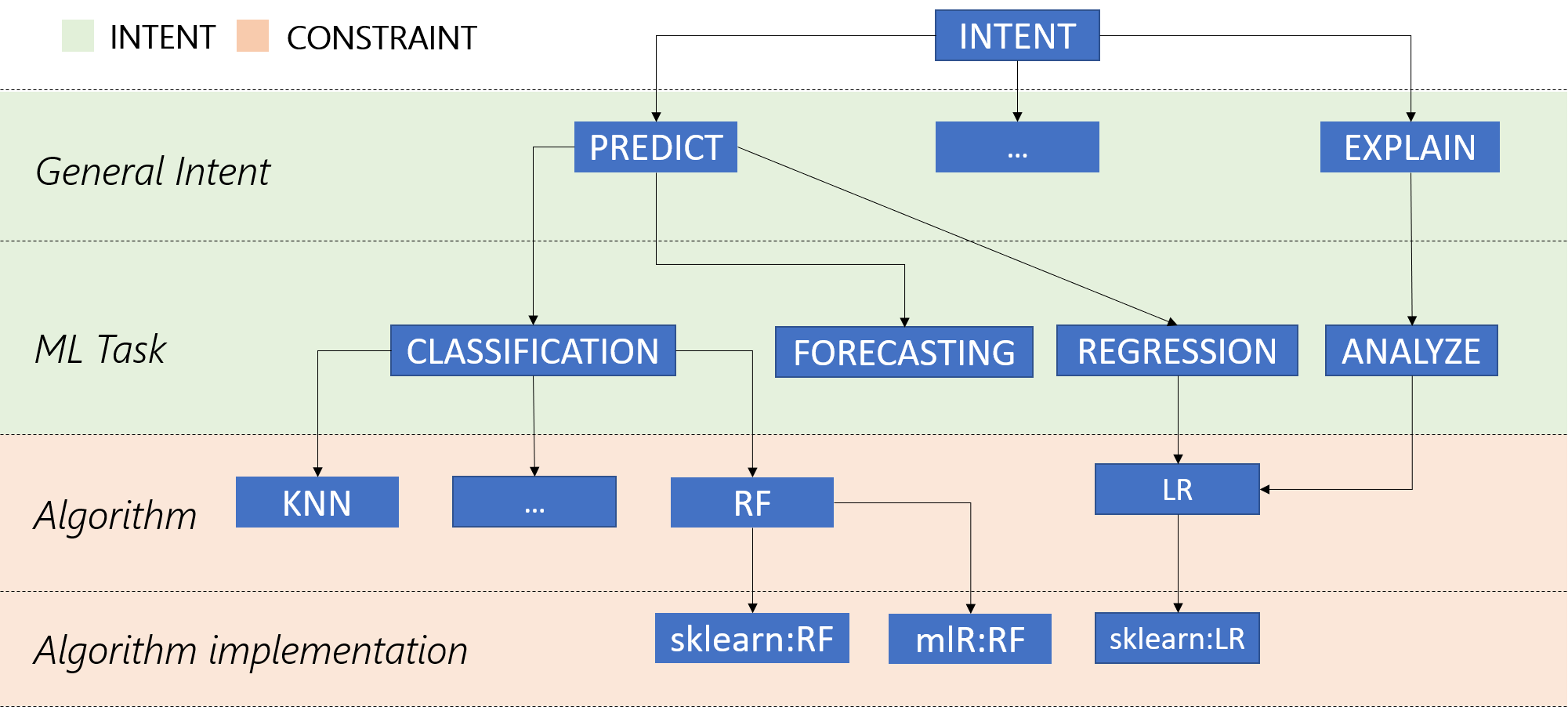}
  \caption{Hierarchy of Intents and Algorithm Constraints.}
  \label{fig:intent}
\end{figure}

In order to support the automated creation of DA workflows, we propose to represent intents within a hierarchical structure that ultimately relates them to concrete pieces of code that can be executed (see Figure~\ref{fig:intent}). Notice that the objective of this hierarchy is to provide a method to represent intents and their relationships, rather than providing a complete ontology for this domain. In this hierarchy, the intents listed above are located in the upper level, while the concrete ML tasks corresponding to them can be found in the subsequent tier. For instance, the \textit{Predict} intent can be broken down into \textit{Classification}, \textit{Regression} and \textit{Forecasting} tasks, the \textit{Explain} intent into \textit{Summarize} and \textit{Analyze} or the \textit{Assess} intent into \textit{Validate} and \textit{Compare}. Following the hierarchical structure, the different ML algorithms can also be paired to the ML tasks they address, and finally concrete implementations for the algorithms are located in the lowest level of the hierarchy. 

It must be noted that a particular ML algorithm can satisfy different Intents depending on its usage. For instance, Linear Regression satisfies the \textit{Predict} intent when the goal is to predict the target variable based on the values of the explanatory variables, but it also satisfies the \textit{Explain} intent when the coefficients given by the algorithm to the different explanatory variables are studied.

\subsubsection{Constraints/Preferences}
The user can state some constraints and preferences on the workflow to be generated. For instance, some constraints could be: 
\begin{itemize}
    \item \textbf{Algorithm Constraints}: the users can specify if they want a specific algorithm to be used or to be dismissed by the generator. As it can be seen in Figure~\ref{fig:intent}, forcing the system to use an algorithm also constrains the possible Intents. 
    \item \textbf{Hyperparameter Constraints}: the user can place constraints on the hyperparameter values. These constraints may require the value to meet specific conditions, such as having a minimum or maximum value or being equal to a predetermined one.
    \item \textbf{Workflow Constraints}: the user can limit, for instance,  the time or memory allocated to workflow optimization.
\end{itemize}

It is clear that the user can pose constraints not only on the elements mentioned above but on every element of a data analytics process. For that, the KG graph has been designed in a flexible manner to enable effectively representing them, as will be described in Section~\ref{sec:kgc}.

\subsubsection{Evaluation Requirements} 
\begin{figure}[t]
    \centering
    \includegraphics[width=0.55\textwidth]{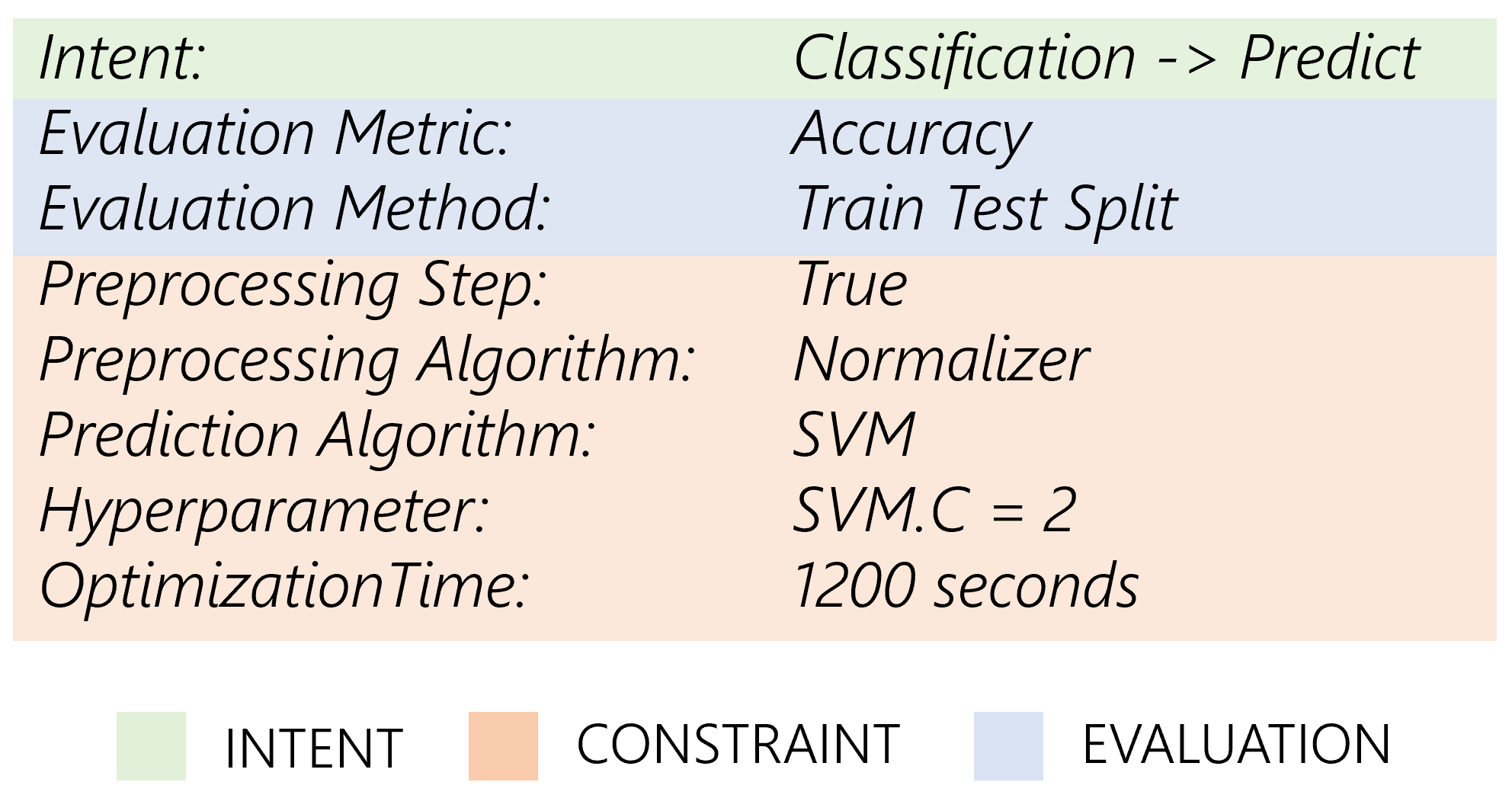}
  \caption[User Input]{Example of User Input.}
  \label{fig:input}
\end{figure}

The evaluation requirements refer to the metrics that the users want to optimize (e.g., minimize, maximize, etc.) and how they want to evaluate them (e.g., cross validation, benchmarking, train-test split, etc.). These metrics will guide the optimization of the workflow and depend on the user intents (e.g., for classification tasks metrics such as accuracy, recall or F1-score are used, while for regression tasks the metrics could be RMSE or MAE).
\newline
\newline
A complete example of an expected user input can be seen in Figure~\ref{fig:input}.

\subsection{Design Methodology}
A KG has been developed to establish a standardized terminology for annotating ML workflows. The methodology used for its creation is based on the one presented in~\cite{ontology101}, whose steps (although modified for this work's characteristics) are explained next.

\subsubsection{Domain and scope}
The KG has been designed to cover all the processes of the creation of ML analytical flows. For that, it should represent the different steps of the workflows, their characteristics and how they are connected. Moreover, it should take the user into account, by annotating their interaction, both the input (e.g., datasets with its characteristics, preferences, etc.) and the output (i.e., feedback), and their context (e.g., user’s education, user’s organization, etc.). The KG should be able to answer general queries about the DA process (e.g., Which preprocessing algorithm is more frequently used before a Random Forest classifier?) but also user specific questions (e.g., Which algorithm constraints has \textit{user-11} previously used with multiclass classification problems?).

\subsubsection{Reusing existing resources}
\label{sec:reuse}
When designing KGs, data engineers are encouraged to utilize existing ones if possible as, besides reducing the effort required to model a domain, reusing existing ontologies enhances interoperability across existing and future applications.

For the design and creation of the proposed KG, two different existing ontologies have been reused. First of all, the well known Person\footnote{https://schema.org/Person} ontology has been reused to define the user. It can capture relevant information, which could be leveraged in the workflow recommendation process (e.g., the users' level of expertise, their interests, etc.).
    \begin{figure}[h]
    \centering
    \includegraphics[scale=0.55]{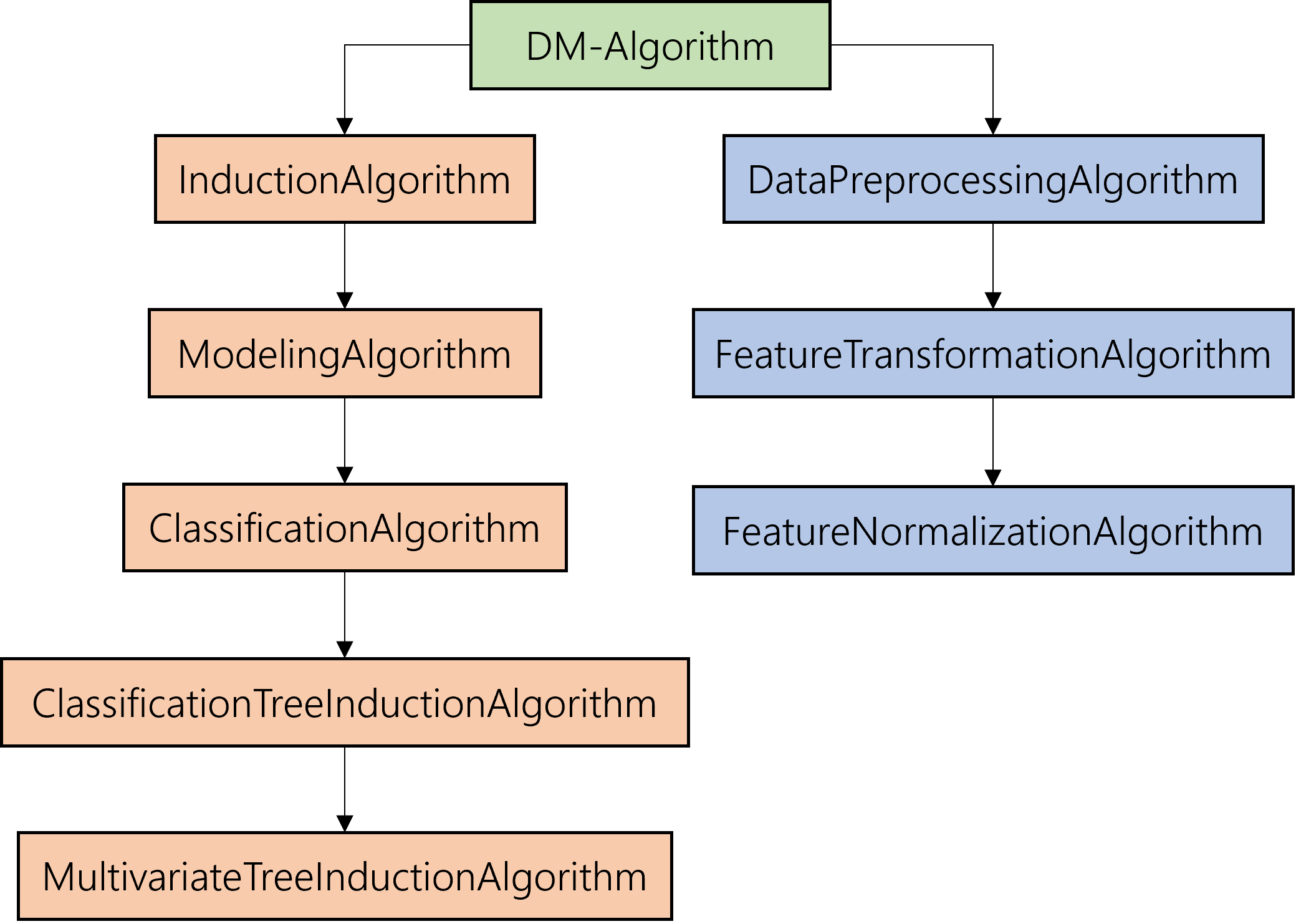}
    \caption[DMOP Hierarchy]{Examples of the DMOP class hierarchies.}
    \label{fig:DMOP}
    \end{figure}

Then, from the DA ontologies presented in Section~\ref{sec:related_work}, DMOP has been chosen for reuse, as it is an ontology designed to support informed decision-making at various points of the DA process. Concretely, it focuses on the automation of algorithm selection by providing annotations of the characteristics of the different workflow steps. This feature makes DMOP stand out among the other options, as although they all model the different parts of analytics workflows, they do not capture the characteristics of the components. Concretely the DMOP's Dataset and Algorithm classes have been reused. The former allows to describe a dataset, with its features and characteristics, while the latter contains many subclasses that classify and annotate a wide range of different algorithms (see Figure~\ref{fig:DMOP}). For instance, the subclass \textit{MultivariateTreeInductiveAlgorithms} (which are instantiated with algorithms like Random Forests) has the property \textit{ToleratesIrrelevantFeatures} and \textit{HandlesCategoricalFeatures}, which can be useful for the automatic creation of pipelines. This ontology is not exhaustive in the algorithms (mainly due to the appearance of new ML techniques over time), hence new classes have been incorporated to it (see Section~\ref{sec:supplementary}). The usage of DMOP influenced the decision of the representation language, which has been OWL~\cite{owl}.

\subsubsection{Knowledge Graph's Contents}
\label{sec:kgc}
\begin{figure*}[t]
  \centering
  \includegraphics[width=\textwidth]{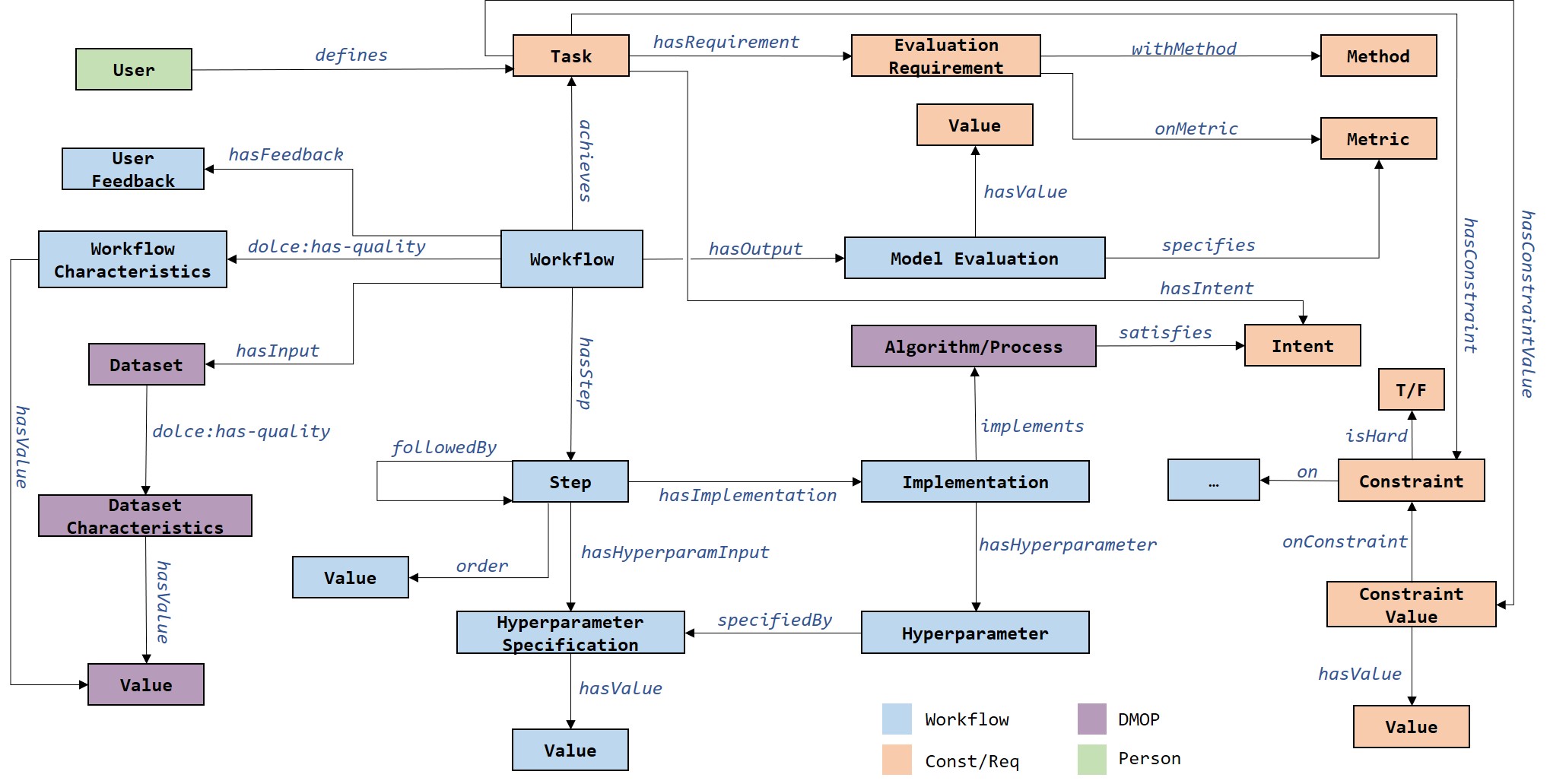}
  \caption[KG Classes and Properties]{Overview of the KG's classes and properties.}
  \label{fig:kg}
\end{figure*}

The central part of the KG (see Figure~\ref{fig:kg}) is the requested task (\textit{Task}), which is defined by a user (\textit{User}), and is linked to the new concepts introduced in Section~\ref{sec:user_input}, namely the user's intent (\textit{Intent}), the evaluation requirements (\textit{Evaluation Requirement}) and the constraints (\textit{Constraint}) that the user defined for the requested task. For the latter, the \textit{isHard} relation has been introduced, which has been created in order to model with the same structure constraints and preferences (i.e., when a constraint entity is linked through an \textit{isHard} relation to True, we are representing a constraint, and a preference when it is linked to False). The task is achieved by a workflow (\textit{Workflow}), which is linked to the data used (\textit{Dataset}), the results (\textit{Model Evaluation}), the user's feedback (\textit{Feedback}) and it's different steps (\textit{Step}). The latter contains information about the algorithms used (\textit{Algorithm}) and their hyperparameters (\textit{Hyperparameter}). The different steps are linked with the \textit{followedBy} relation, which has been modeled without cardinality restrictions, hence a particular step can be linked to multiple others (either proceeding or succeeding it) enabling the representation of complex and non-linear workflows.

Regarding the reused classes (see the green and purple classes in Figure~\ref{fig:kg}), the \textit{User} class has been defined by mapping it to the \textit{Person} class in the Person ontology. Likewise, the \textit{Dataset} and \textit{Algorithm} classes have respectively been mapped to the \textit{DM-Dataset} and \textit{DM-Algorithm} classes in the DMOP ontology. Additionally, the superclass \textit{DatasetCharacteristics} and its corresponding subclasses have also been reused and extended to incorporate the desired characteristics (see Table~\ref{tab:features}).  

As will be seen in Section~\ref{sec:workflow_instantiation}, the proposed KG is able to capture all the workflows generated by popular AutoML tools.
\section{Instantiating the Knowledge Graph}
\label{sec:population}
The KG presented in Section~\ref{sec:kg_design} has been designed with the objective to annotate the end-to-end interaction of a user and the workflows generated within the context of a DA Assistant. This is, once a user requests a ML task, the inputs, the generated workflow and the feedback should be automatically annotated following the defined KG. This information will then be used by the system to produce new recommendations once a new pipeline is requested by the same or another user. Therefore, this implies that the KG would have no data until the system is operational, and would suffer from the cold start problem (i.e.,  the difficulty of making accurate predictions or recommendations for new or unseen data points with limited or no prior information). 

Hence, in order to overcome these difficulties, data regarding the various components that constitute a ML workflow (see Section~\ref{sec:supplementary}), along with  workflows themselves (see Section~\ref{sec:workflow_instantiation}), have been obtained and annotated in the KG.

\subsection{Data Instantiation}
\label{sec:supplementary}
To create instances for the different DA elements (e.g., algorithms, datasets, hyperparemeters, etc.), various resources have been used. First of all, DMOP’s knowledge base DMKB~\cite{dmkb} has been reused. There, different algorithms are annotated by linking them to the algorithm class they belong to (thus inheriting the class’ properties) and also by stating their specific characteristics. For instance, \textit{Logistic Regression} is instantiated in DMKB as a \textit{Discriminative Algorithm}, with its characteristics (e.g., assumptions, optimization strategy, output generated, etc.) and its three main properties, namely \textit{EagerPolicyLearning}, \textit{HandlesBinaryClassification}, and \textit{HandlesContinuousFeatures}. Moreover, by being a \textit{Discriminative Algorithm}, the algorithm also inherits the \textit{Discriminative Algorithm}'s properties. All of the available algorithm classes in DMKB have been linked to the intent they satisfy down to the ML Task level (see Figure~\ref{fig:intent}). However, DMKB does not contain all the ML algorithms that currently exist, hence it has to be extended with the ones that the system will use. To this end, instances of the algorithms need to be created, linking them to the intent they fulfill, to the corresponding algorithm class in DMOP's hierarchy and to their hyperparameters. In this work, the workflow generators (see Section~\ref{sec:workflow_instantiation}) use the popular Python library Scikit-learn~\cite{scikit-learn}, hence its components have been instantiated in the KG.

Regarding the datasets, different repositories have been used depending on the analytical purpose they address. For classification problems, data has been extracted from a collection of 68 datasets from OpenML~\cite{openml}, and for regression problems, 28 datasets have been extracted from the UCI repository~\cite{uci}. For all the datasets, simple but relevant characteristics have been automatically extracted and annotated, which are summarized in Table~\ref{tab:features}.

\begin{table}[htpb]
\caption[Dataset Features]{Example of features extracted from the datasets, depending on the target variable.}
\centering
\begin{tabular}{ccc}
\toprule
 & \textbf{Categorical} & \textbf{Numerical} \\ 
 \midrule
\# of Instances & \cmark & \cmark \\ 
\# of Features & \cmark & \cmark \\
\# of Numeric Features & \cmark & \cmark \\ 
\# of Categorical Features & \cmark & \cmark \\ 
\% of Missing Values & \cmark & \cmark \\
\# of Classes in Target & \cmark &  \\ 
Target Imbalance & \cmark &  \\ 
StD Target &  & \cmark \\ 
Target Type & \cmark & \cmark \\ 
\bottomrule
\end{tabular}
\label{tab:features}
\end{table}

\subsection{Workflow Annotation}
\label{sec:workflow_instantiation}
Ideally, annotated workflows should have been obtained from users' interactions with DA Assistants. However, as such datasets do not exist, instances have been obtained by emulating the generation of different optimized workflows. Therefore, the KG has been populated by annotating the results of different AutoML tools, which have been used to optimize pipelines for the datasets from OpenML and from the UCI repository, using the elements instantiated in the previous section. This is, multiple workflows have been optimized for each dataset, with different evaluation metrics and constraints, and have been annotated according to the classes and properties shown in Figure~\ref{fig:kg}. Concretely, the tools used (which are both implemented in Python and use the ML library scikit-learn) are HyperOpt-Sklearn~\cite{hyperopt} and TPOT~\cite{tpot}. They are AutoML tools which optimize ML workflows by choosing algorithms and their hyperparameters. They use different search strategies (e.g., Tree of Parzen Estimators, Genetic Programming, etc.) and return pipelines which consist of optional preprocessing steps and a prediction algorithm (classification or regression). They have the ability to constrain the search space by choosing the algorithms, fixing their hyperparameters, imposing pipeline templates, or limiting the time or number of steps allocated to the optimization process. 

The expressivity of the KG allows to completely annotate the input (e.g., all the available constraints, the datasets, etc.) and the workflows generated by these tools. For instance, the template constraint \textit{Selector + Classifier} could be annotated as shown in Figure~\ref{fig:template}. However, even though these tools do not offer all the possibilities represented in the KG, they have not been extended as the creation of a workflow generator is out of the scope of this work. 

\begin{figure}[tp]
    \centering
    \caption{Instantiation of a template constraint.}
    \label{fig:template}
    \end{figure}
\section{Anticipating User Intents}
\label{sec:anicipation}
The KG designed so far solves the problem of capturing the whole analytics workflow, including information about the users. Next, we explain how the information stored in the KG can be leveraged to assist the users when they are creating  analytical workflows. To this end, we provide recommendations for their input (see Figure~\ref{fig:general} arrow 2), which can be crucial for generating relevant workflows. For instance, if the users select a dataset, suggestions of potential analyses based on the dataset characteristics should be provided to them. It is important to note that these user inputs, enumerated in Section~\ref{sec:user_input}, are subject to the user's objective, hence there does not exist a general correct answer. For instance, given a task that is well known to obtain the best results when approached with Neural Networks, a user could constrain the algorithm to a Decision Tree if their current goal is to create an interpretable model. In the case of datasets, a regression dataset could be given the classification Intent if the user plans to discretize the response variable. Therefore, the goal is to produce sensible recommendations based on the available information, which the users may decide to use or not.

To achieve the anticipation objective, a basic approach based on the exploitation of predefined graph queries has been explored first, and then its main limitations have been addressed by taking advantage of the whole KG structure by using knowledge graph embeddings. 

\subsection{Basic Approach}
\label{sec:queries}
The first proposed method to exploit the KG involves using predefined SPARQL queries to retrieve the information needed to assist users. These queries can be defined to take advantage of past interactions with the system, both from the user requesting the experiment and from others. Therefore, it is a highly interpretable method, designed with domain knowledge heuristics, which can overcome the lack of user’s interaction data by giving relevance to the expert creating the queries. Moreover, contrarily to most recommendation algorithms, it can automatically exploit new information to provide the recommendations, as there is no model to be retrained. 

\begin{figure}[htpb]
    \centering
    \includegraphics[width=0.8\textwidth]{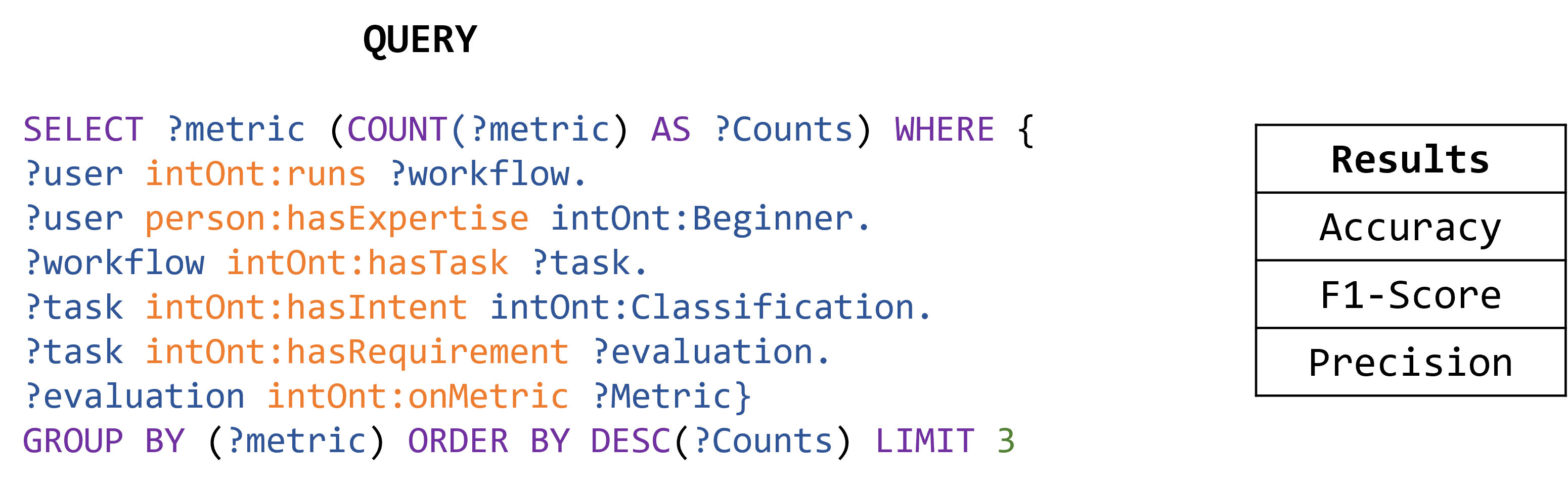}
  \caption{Example of an SPARQL query over the KG.}
  \label{fig:query}
\end{figure}

To illustrate the approach, let's think of a user who requests a workflow for the Iris dataset and his intent is classification. Then, the system could suggest the metric for which the workflow generation should be optimized. To answer it, the system starts by running a very specific query, and it keeps increasing the level of generalization until an answer is returned. For instance, the system could run \textit{Query 1: Which is the most frequent metric the user has specified in the past with the Iris dataset for a Classification task?} However, the user may not have used the Iris dataset before, hence another query could be used to allow more generalization: \textit{Query 2: Which is the most frequent metric the user has specified in the past for Classification tasks?} Yet, it may be the first time the user interacts with the system, hence different queries must be run in this case. \textit{Query 3: Which is the most frequent metric used with the Iris dataset for a Classification task by all the users?} in case the Iris dataset has been used before, or \textit{Query 4: Which is the most frequent metric used in Classification tasks by all users?} when the Iris dataset has not been used. These queries can be further restricted with different filters, such as the domain of the analysis or the size of the dataset. In Figure~\ref{fig:query}, a SPARQL example of Query 4 filtered by the users' level of expertise is shown.

Therefore, different sets of ordered query templates can be defined to provide suggestions depending on the information available or on the intended level of generalization. These sets can be found in the prototype\footnote{https://github.com/gerardponsrecasens/capturing-and-anticipating-ML-intents} presented in Section~\ref{sec:prototype}.

We have seen how users can be assisted by retrieving information from the KG with different SPARQL query templates. The suggestions produced by these queries are sensible and highly interpretable, as they should be defined by domain experts, and allow using new information by just storing it in the KG. These queries are useful in the situations where there exist enough data in the KG about the user, the dataset or the task, which allows to return concrete recommendations.

However, the query definition poses some limitations. First of all, it requires some decisions to be made in advance (e.g., the order of the queries). Additionally, depending on the level of generalization, the definition of these queries could become complex and inflexible, especially if information about the whole KG is needed (e.g., suggesting an evaluation metric based on similarities between datasets).

\subsection{Exploiting the Graph's Structure}
\label{sec:embeddings}
Besides offering a very powerful way of storing and representing data, the role of graphs is not only restricted to their use as knowledge bases. For instance, ML techniques can be applied to KGs, which can be used to address the limitations posed by the querying method. These techniques range from community detection to link prediction, and can be beneficial in a lot of applications. However, the more advanced analytical methods suffer either from high computational costs if they run directly on the graph, or from the inability to be used on top of graph structures. To deal with these problems, the information stored in the graph can be represented in fixed length vectors in a low-dimensional vector space, which can then be fed naturally into ML algorithms. These transformations are known as knowledge graph embeddings.
\subsubsection{Knowledge Graph Embeddings}
\begin{figure}[htpb]
\centering
    \includegraphics[width=0.8\textwidth]{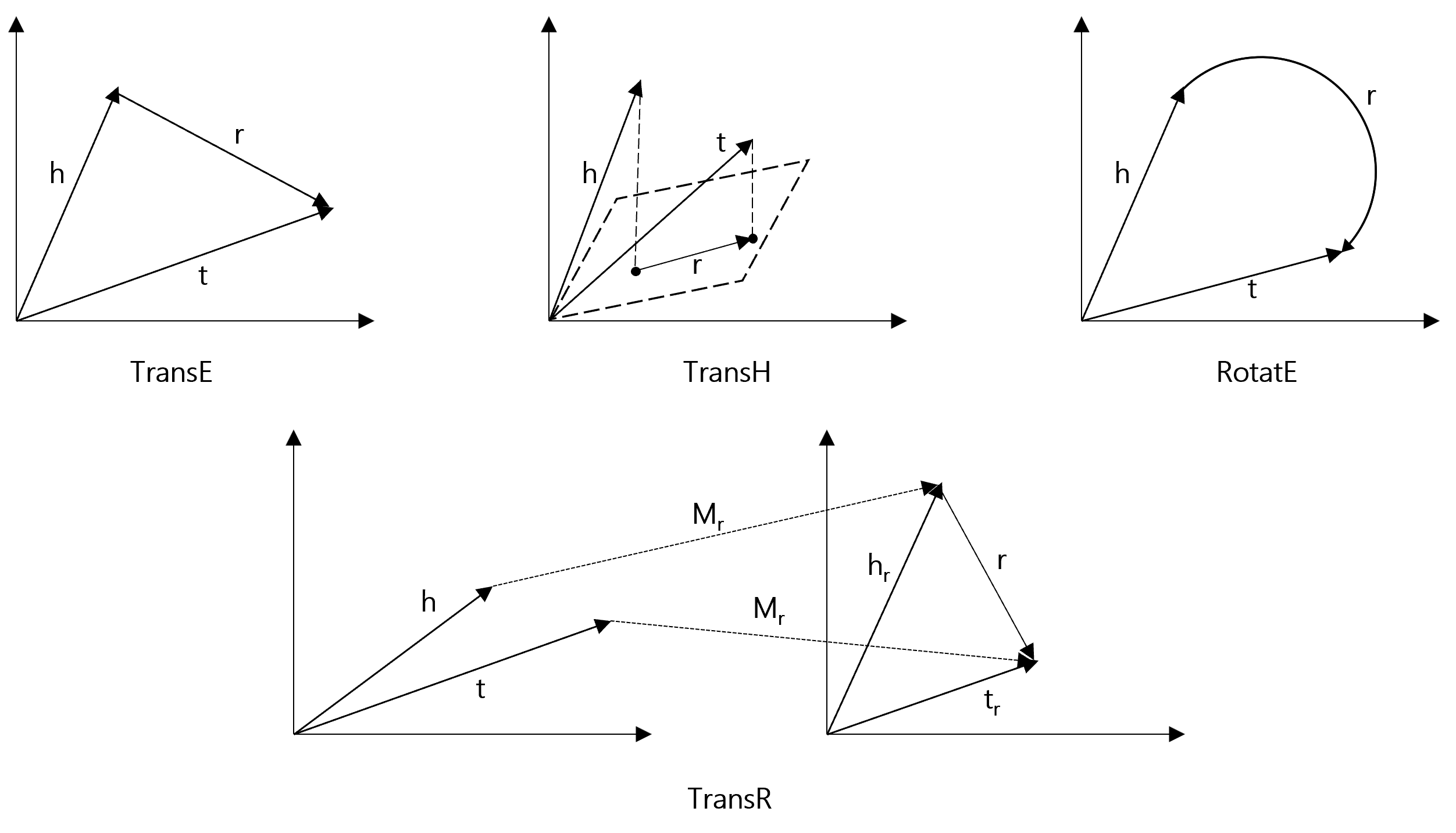}
  \caption{Simplified 2D visualization of the intuition behind the different translational embedding models.}
  \label{fig:translational}
\end{figure}

 Knowledge graph embeddings (KGEs) are transformations of the different elements (i.e., entities and relations) of a KG into low-dimensional feature vectors, which focus on preserving as much original information from the graph as possible. There are multiple techniques to obtain these KGEs representations. Translational models (see Figure~\ref{fig:translational}) represent the relationships between entities by translating the embedding of the head entity to the tail entity using the relation entity, with methods such as TransE~\cite{transe}, TransH~\cite{transh}, TransR~\cite{transr} or RotatE~\cite{rotate}. Bilinear models optimize the bilinear product between the entities of a relationship, with algorithms such as DistMult~\cite{distmult} or ComplEx~\cite{complex}. More recently, methods exploiting Neural Network based approaches have also been considered, such as ConvKB~\cite{convkb}.

Independently of the technique, these generated KGEs can be used for the anticipation task by creating ML algorithms able to recommend or suggest the inputs to the users. These algorithms are known as Recommender Systems (RSs), which are tools that aim to provide personalized \textit{item} recommendations to \textit{users} based on their preferences, past interactions and/or item characteristics. KGs are an ideal data structure for RSs~\cite{recommender,recommender1}, as they naturally capture the user-item interactions as well as the user and item properties. Therefore, KGE could enrich the elements' representations  by capturing complex characteristics, such as relationships or context. Moreover, the embeddings allow to partially mitigate the cold-start problem for the recommendation of new artifacts, as they benefit from being linked to already embedded entities and relations. For instance, a new algorithm will have an associated Intent and will be linked to some algorithm classes and characteristics from DMOP, which will be shared with other algorithms already present in the KG. Therefore, these algorithms will be close in the embedding space, which can be leveraged for the generation of recommendations.

However, the anticipation needs for the user intentions modeled in this work have multiple interactions to be recommended, resulting in different objects acting as users or as items which in turn evolve while the user’s input is being extended. To exemplify the problem, we could think about a user who starts requesting a workflow. First of all, the system could suggest to them the dataset to be used. This first interaction can be clearly recommended, following the terminology presented above, with a \textit{user:User} and \textit{item:Dataset} RS. Once the dataset is fixed, the system should provide anticipations for the Intent. There, if a similar RS was to be used (\textit{user:User, item:Intent}), the information of the selected Dataset would be lost. Hence, a RS with \textit{user:(User+Dataset)} and \textit{item:Intent} should be used instead. The problem gets worse once the number of inputs increases or if the difference in the order of the inputs (e.g., the user could select first the Intent and expect the system to provide appropriate datasets) is considered, making the use of simple RS not feasible.

\subsubsection{Link Prediction} 
\begin{figure}[htpb]
    \centering
    \includegraphics[width=0.8\textwidth]{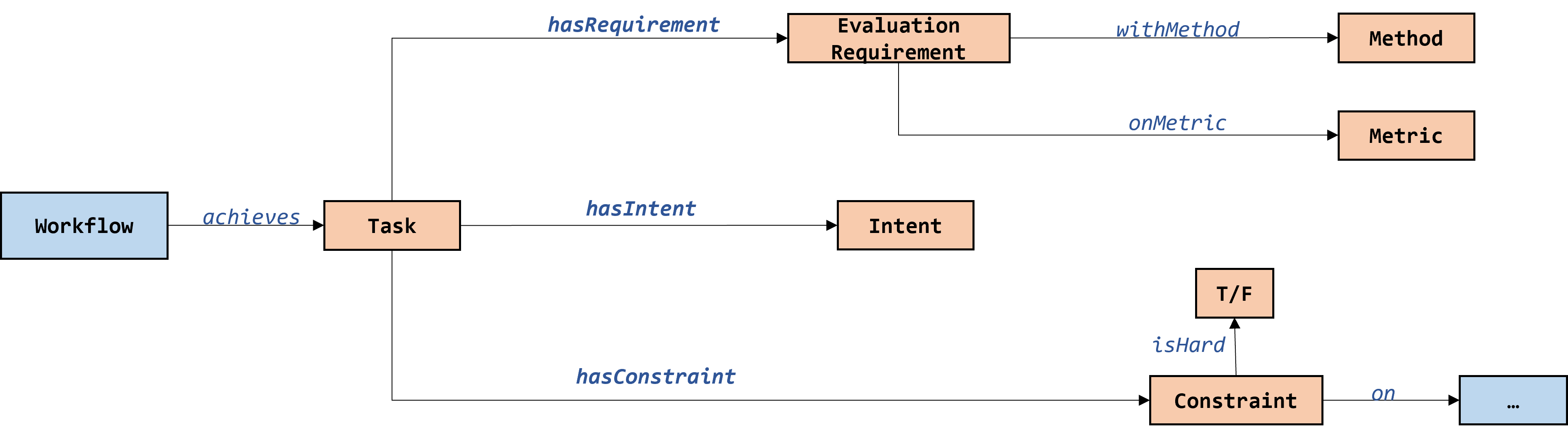}
  \caption[Embedding Dimension]{Overview of classes and properties relevant for the anticipation task.}
  \label{fig:kg_filtered}
\end{figure}

To overcome this difficulty, the Link Prediction (LP) method has been explored to act as a RS. LP is a technique that can be applied over KGs ~\cite{lp,lp1} in order to find missing or future relations between entities. Therefore,  given that the users’ inputs have been captured in the designed schema (i.e., all the relations needed for the anticipation task are present in the KG), the KG structure can be exploited with the LP task to provide suggestions. This procedure can be done with different methods, such as measuring similarity between entities or by using the graph topology (e.g., Common Neighbours). KGE can be used to predict these missing links as well, by  using the model trained for the embeddings. For instance, if the model is trained to minimize translation distance by addition like in TransE (i.e., $h+r \approx t$, see Figure~\ref{fig:translational}), we can look for the entities whose tail embeddings are closer to the summation of the head and relation embeddings. Thus, this technique gives flexibility to the recommendation engine, as it does neither lock the input nor the output of the recommendation. The relations that can be used to anticipate the user’s input (i.e., the elements that define the task the workflow aims to solve) are highlighted in Figure~\ref{fig:kg_filtered}. The general intention of the requested workflow is captured by \textit{hasIntent}, \textit{hasRequirement} captures the evaluation protocol and \textit{hasConstraint} links the workflow with the different constraints the user may pose.

\section{Experimental Evaluation}
\label{sec:experiments}
In this section, different experiments of the application of KGEs and LP to the designed KG have been performed, having two main goals. On the one hand, they aim to evaluate the KGEs' ability to capture the whole KG, by comparing different embedding models and parameter configurations. On the other hand, they assess to which extent the LP task can be used to provide sensible recommendations in the studied domain.

\subsection{Methodology}
The experiments have been created following the general methodology presented in~\cite{performance}, which focuses on good practices in the experiment design for computer systems.

\subsubsection{Goals and Systems} The experiments have been conducted to evaluate the performance of different KGEs in order to determine their ability to capture the KG and to potentially be effective for the general LP task. While doing so, the influence of the different parameters on the performance has been studied.
\subsubsection{Services and Outcomes} To produce the experiments, the PyKeen~\cite{pykeen} Python library has been used. This library enables the creation of reproducible KGEs experiments, with multiple functionalities, tuning parameters and models. With that, from a set of triples, KGEs have been learned in a training phase and assessed over a validation set in the evaluation stage.

\subsubsection{Metrics} As in the vast majority of ML tasks, in KGE a loss function is optimized in the training phase to obtain the most suitable embeddings. When KGEs are learned for LP, one of the most usual loss functions is the Margin Ranking Loss (MRL):

\begin{equation}
  MRL = \sum_{(h_i,r_i,t_i)}max(0,m+f(h_{i}^+,r_i,t_i)-f(h_{i}^-,r_i,t_i)) 
  \label{eq:mrl}
\end{equation}
The objective of this loss function is to give a better score to positive triples (i.e., those that exist in the dataset) than to the negative ones (i.e., those which are synthetically generated by corrupting some of the entities of the positive triple~\cite{negative}), emulating the LP task. When the difference between the score of the two entities is larger than the specified margin (\textit{m}), no loss is accounted for. The score function \textit{f} depends on the given embedding model. For instance, for TransE, which is a simple embedding technique that sees the translation as the addition of the head and relation embeddings, the scoring function is the q-norm of the mismatch of the translation in the embedding space: 
\begin{equation}
    f(h,r,t) = -\left \| e_h+e_r-e_t \right \|_q
\end{equation}

During the evaluation phase, the head $(?,r,t)$ and the tail $(h,r,?)$ are removed from the triple of the test set, and LP is performed on both incomplete triples. 

The Hits@k metric is usually used to assess the evaluation, as it measures the proportion of times the correct result (i.e., the entity missing for the LP) is shown among the top-k ranked entities. Therefore, it replicates the case of showing only a few highly ranked items to users in recommendation engines. Moreover, given the expected usage of the LP task (i.e., present the users some suggestions), the evaluation metric that has been given more relevance is Hits@3. This is because using three suggestions presents the users a sufficient amount of choices while not overwhelming them with excessive options.

\subsubsection{Parameters}
The main parameters that can affect the generation of the embeddings are:
\begin{itemize}
    \item \textbf{Models}: Different proposed models capture the head and tail relationship based on different principles and assumptions. Hence, the performance on the LP task can be widely influenced by the model choice. 
    \item \textbf{Embedding Dimension}: This is the dimension of the vector space in which the entities are represented. In general, the larger the dimension, the more hidden relationships that can be captured by the model. However, increasing the dimension can lead to overfitting on the training data (and thus poorer performance over unseen triples) and computational complexity problems. Therefore, a wide range of values will be explored to find a trade-off between the complexity of the hidden relationships captured and generalization.
    \item \textbf{Learning Rate}: The learning rate is an important hyperparameter in the training of various ML algorithms, as it determines the step size at which the model updates its parameters during the training process. It can have a significant impact on the performance of the model, as a high learning rate could cause the model to overshoot the optimal solution, leading to unstable behavior, or, on the other hand, a low learning rate could result in slow convergence or the model getting stuck in a local minimum. 
    \item \textbf{Number of negative samples per positive sample (NPP)}: Previous studies~\cite{negadiff} indicate that increasing its value may improve the evaluation metric, albeit at the cost of greater computational complexity. This is because by increasing the NPP the model learns to discriminate better the negative samples, thus generating less False Positives in the inference phase. However, increasing the value too much could hinder the model’s ability to detect positive instances, which greatly affects the LP task. 
    \item \textbf{Negative Samples Generator}: Different negative samples generators can be used (e.g., Uniform Sampler, Bernoulli Sampler, etc.), which can influence the results by creating training samples of different quality.
    \item \textbf{Number of Epochs}: The number of epochs of the experiments can also affect the results. For instance, a low number of epochs can lead to underfitting, as the model has not had enough time to learn about the data, but a large number of them could lead to overfitting if the model is able to memorize the training data and lose generalization. 
\end{itemize}

\subsubsection{Factors to Study}
 In the experiments, the variables that have been studied are:
\begin{itemize}
    \item \textbf{Models}: Popular translational (i.e., TransH, TransE, TransR, RotatE) and bilinear models (i.e., DistMult, ComplEx) have been studied. 
    \item \textbf{Embedding Dimension}: The values tested range from 2 to 256, in the powers of two, as it is a usual rule-of-thumb in structure sizes when GPUs are used. In all our experiments the performance of the model stabilizes with embedding sizes within the proposed range.
    \item \textbf{Learning Rate}: The range of values explored starts at 0.1, and then exponentially lower values (e.g., 0.01, 0.001) have been assessed.
    \item \textbf{NPP}: The explored range has been between 1 (the usual default value) and 100.
\end{itemize}

The remaining two variables (i.e., \textit{Negative Samples Generator} and \textit{Number of Epochs}) have not been studied as they have been fixed in advance. First of all, for the \textit{Negative Samples Generator}, the Bernoulli generator has been used instead of the Uniform one, as it generates higher quality corrupted triples for the training phase by selecting the best entity (i.e., tail or head) to change. Regarding the \textit{Number of Epochs}, although in the experimental set up Early Stopping has been enabled, the maximum value has been fixed to 300, which has not been reached by any experiment.
\subsubsection{Evaluation Techniques}
\label{sec:eval_techiniques}
As previously mentioned, the PyKEEN library has been used to generate and evaluate the experiments. The different combinations of models and hyperparameters have first been tested in a grid search approach. To perform the experiments, the data has been split in three sets (i.e., train, test and validation) in a stratified manner. This is, they have been split taking into account that, in order to perform the test and validation phases, the entities and relationships appearing in their respective sets must have been first embedded in the training phase. To be more specific, the different sets which have been used for each task are:
\begin{itemize}
    \item \textit{Training Set}: This is the dataset partition that is used for learning the embeddings. The different models with the specified hyperparameter configuration are trained with this set by minimizing the MRL function. It is important to note that this loss function can neither be compared between models, as they use different techniques to define distances or errors, nor between parameters (e.g., larger embedding sizes can cause the errors to be higher due to vector dimensionality, or a higher number of negative samples per positive sample causes the summation on the loss function to contain more terms). Hence, the results of the loss of the training phase have only been used to assess the performance in scenarios where all the other variables that influence it are fixed. 
    \item \textit{Validation Set}: In order to use the correct number of epochs and to decrease the computational burden of the experiments, the embeddings have been trained with Early Stopping using the validation set. In this experimental setup, the maximum number of epochs has been set to 300, the number of epochs between validation steps to 15 and the patience to 2. With this configuration, if for an amount of time equivalent to 10\% of the total time allocated for training the model has not improved, the learning is stopped. 
    \item \textit{Test Set}: After the training has finished, the final embeddings are used to evaluate the test partition. With that, the different metrics (e.g., Hits@k, average rank, etc.) are obtained. These results have been used to compare the different models and configurations. 

\end{itemize}

One thing to note is that with the standard LP evaluation, all the entities present in the KG are ranked and used in the Hits@k metrics. Hence, each of the thousands of entities could be presented as a solution, which is not the case of the envisioned anticipation system, which should only suggest plausible entities. This is, the only ranked entities should be those whose class satisfies the range condition of the relation, using the ranking score to disambiguate between them. Moreover, as the users introduce their input in an incremental approach, the search space of recommendations gets gradually constrained as new inputs are fixed. For instance, when recommending algorithm constraints, if a user has already chosen an Intent, the LP engine should only rank the algorithms that satisfy it. Thus, the results obtained in these experiments are pessimistic in terms of the entities suggested.

Furthermore, it is worth noting that although in the evaluation phase each triple $(h,r,t)$ is converted to the incomplete triples $(?,r,t)$ and $(h,r,?)$, they should have different importance when considering the results, in concordance with how the LP task will be used in the anticipation phase and with the nature of the KG design. To exemplify it, let’s think about a user, \textit{User10}, who wants to create a DA workflow over a dataset. If we are interested in anticipating the relation \textit{hasIntent}, the system will request the highest ranking entities that satisfy the triple \textit{(TaskU10,hasIntent,?)}, which is one of the incomplete triples generated in the evaluation $(h,t,?)$. It can be seen that using the other incomplete triple \textit{(?,hasIntent,Classification)} adds unnecessary difficulty to the evaluation, as there are multiple workflows that could satisfy that, and does not relate to the anticipation capabilities needed. 
\subsubsection{Dataset}
The dataset used for the generation of the embeddings is made of the triples defining the KG's schema, the triples defining the experiments (both classification and regression), the triples describing the datasets and the auxiliary triples created to instantiate the knowledge base with the scikit-learn concepts used by the workflow generator. This corresponds to more than 20.000 triples, summarized in Table~\ref{tab:data}, which have been partitioned in the three previously defined sets.

\begin{table}[htpb]
\caption{Summary of the number of triples present in each source.}
\centering
\begin{tabular}{l c}
\toprule
 & \textbf{Number of Triples} \\
 \midrule
\textbf{KG schema} & 3252 \\
\textbf{Meta Features} & 2822 \\
\textbf{Scikit-Learn Concepts} & 1285 \\
\textbf{Experiments} & 13262 \\
\bottomrule
\end{tabular}
\label{tab:data}
\end{table}

\subsubsection{Experiment}
The experiments have been distributed in two phases in order to maximize the information without excessive computational resources. On the first one, all the mentioned models have been evaluated with a not extensive number of configurations, in order to identify the general effects of the parameters on the LP performance. When needed, more concrete experiments have been carried out with a wider range of configurations. After that, the most promising KGE model has been chosen and it has gone through an automatic fine-tuning step to get a more optimal configuration of parameters. 


\subsection{Results}
First of all, the general metric used to evaluate the experiments has been Hits@3, due to its resemblance to the recommendation task. Additionally, as explained in Section~\ref{sec:eval_techiniques}, the embeddings are expected to perform significantly better in the tail predictions than in the head ones. This behavior has been clearly observed, as the tail to head average performance value is 8.2 for ComplEx, 2.7 for DistMult, 4.3 for RotatE, 8.6 for TransE, 7.7 for TransH and 9.1 for TransR. Hence, unless a different evaluation is stated, the following experiments, which explore the influence of the parameters under study on the evaluation score,  have been reported with the tail prediction of the realistic (i.e., giving the average ranking position between the entities that have the same loss) Hits@3 metric. 

\begin{figure}[htpb]
    \centering
    \includegraphics[width=\textwidth]{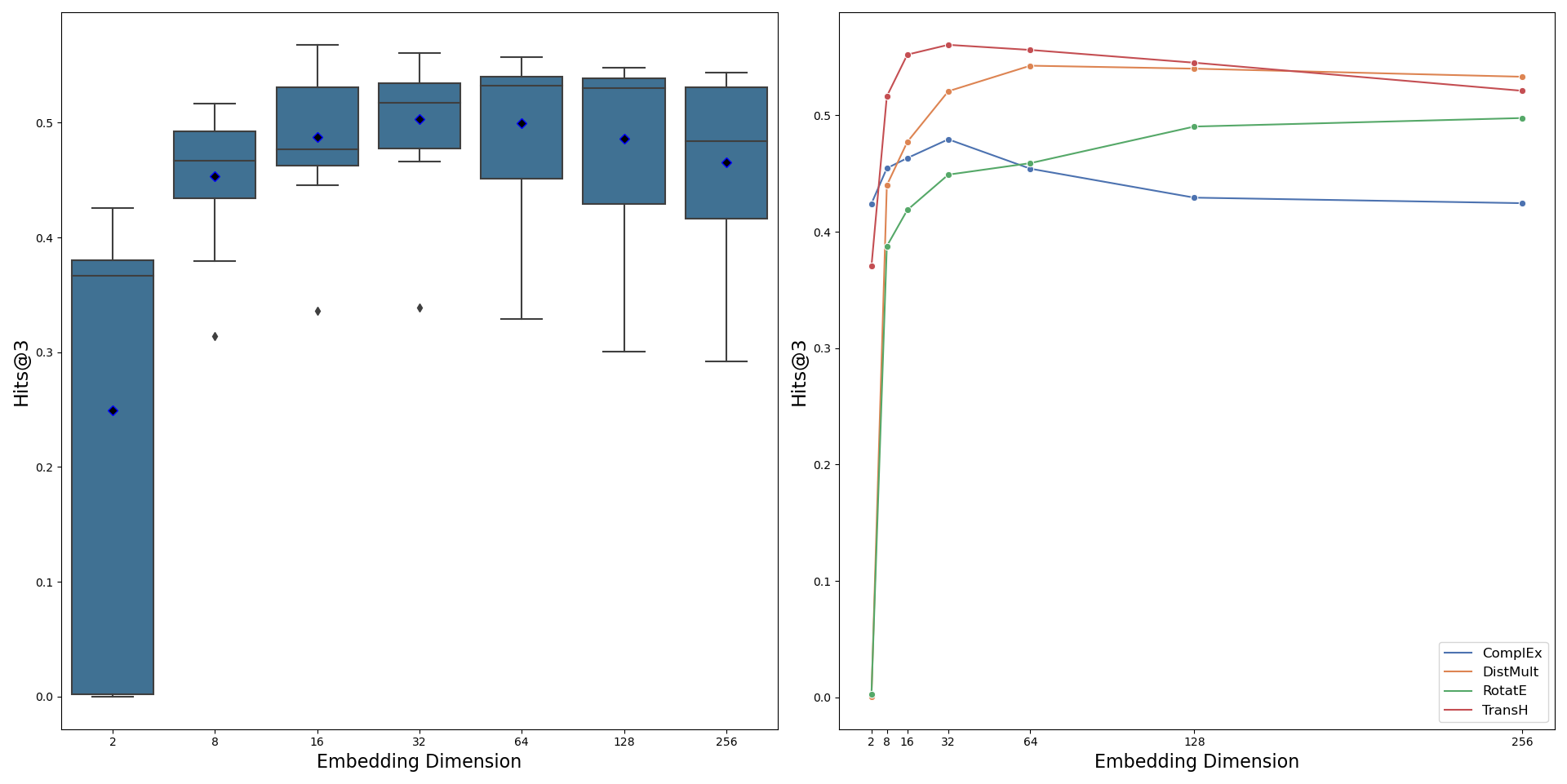}
  \caption[Embedding Dimension]{Hits@3 score grouped by Embedding Dimension (left) and dependence of Hits@3 score with Embedding Dimension for different models, with fixed LR, $0.001$, and NPP $20$ (right).}
  \label{fig:result_embedding}
\end{figure}
For the embedding dimension parameter, the expected and previously described behaviour can be seen in Figure~\ref{fig:result_embedding} (left). The average evaluation performance grows with the embedding dimension size, but it stabilizes (even decreases) for the high values of the dimension (i.e., 128 and 256). This behavior is represented for a fixed configuration in Figure~\ref{fig:result_embedding} (right), where except for the RotatE model, the other models reach their best performance when the embedding dimension presents intermediate values (i.e., 32 or 64). It can also be seen that depending on the model, the embeddings are not capable of capturing the information contained in the KG when a small embedding size (i.e., 2) is used, giving an evaluation score close to 0.

\begin{figure}[htpb]
    \centering
    \includegraphics[width=\textwidth]{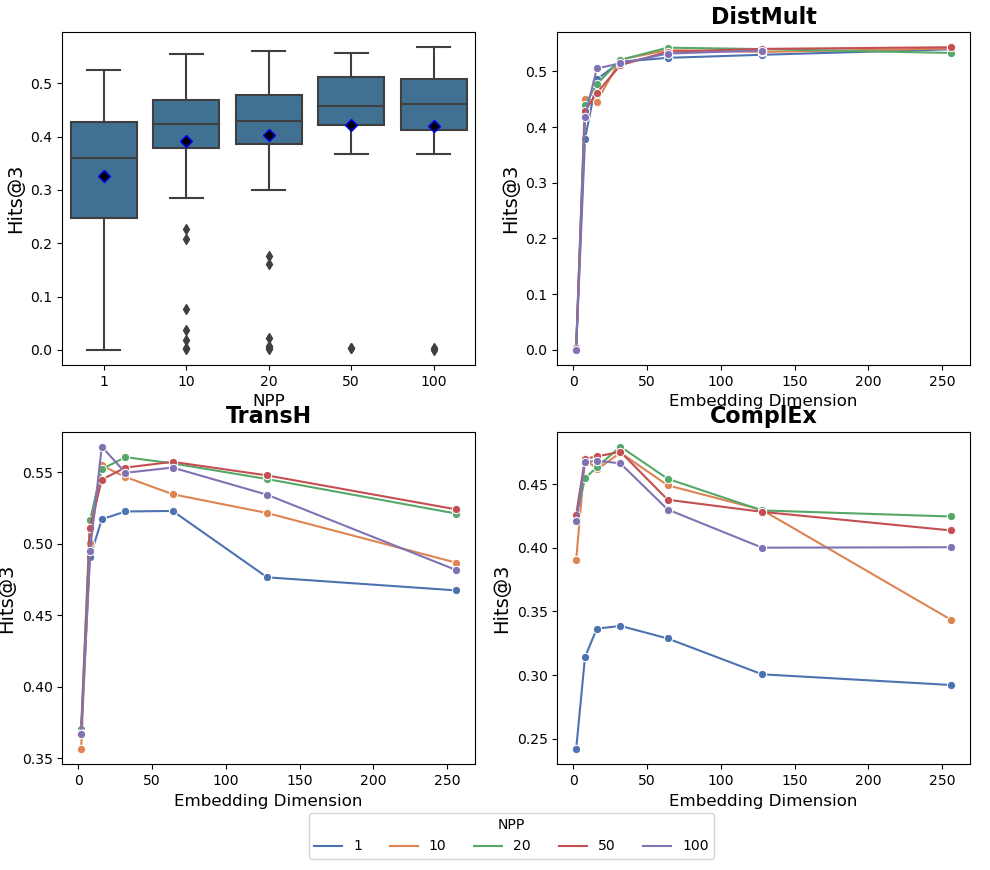}
  \caption[Negatives Per Positive]{Hits@3 score grouped by NPP (top-left) and influence of the NPP parameter on different models.}
  \label{fig:result_npp}
\end{figure}
Regarding NPP, a similar behavior is observed (Figure~\ref{fig:result_npp}, top-left), where there is a stabilization of evaluation score gain as the NPP increases. Hence, it is clear that for the addressed task, using the standard configuration of 1 NPP is not enough, and it is beneficial to perform longer training sessions. However, it is worth noting that this behavior does not apply to all models, as it depends on their scoring function~\cite{negamath}. As can be observed in Figure~\ref{fig:result_npp}, contrary to TransH or ComplEx, which benefit from having more than one NPP, the results in DistMult are not greatly influenced by the NPP.

\begin{figure}[htpb]
    \centering  
    \includegraphics[width=\textwidth]{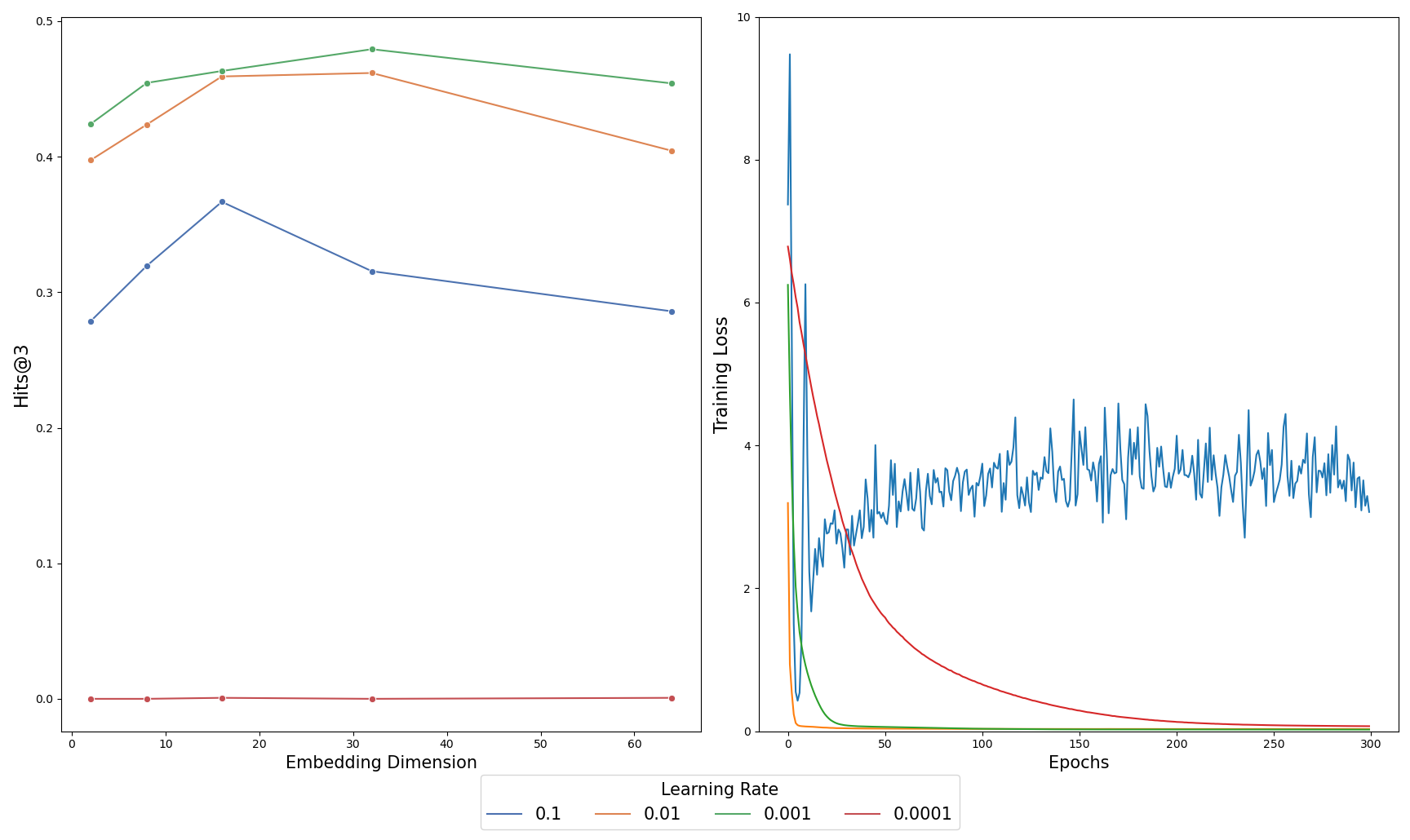}
  \caption[Learning Rate]{Effect of the learning rate on the evaluation (left) and on the training loss for (right) a fixed model (ComplEx, Embedding Dimension=32, NPP=20).}
  \label{fig:result_lr}
\end{figure}

The last parameter is the Learning Rate, whose influence can be easily seen in the evolution of the loss function during the training phase. For that, Early Stopping has been deactivated, so the results can be clearly appreciated (Figure~\ref{fig:result_lr}, right). For high values of the Learning Rate (i.e., 0.1), the learning stage is not able to converge, and it ends with an oscillating behavior. For the rest of values, the loss function reaches values close to 0, with a convergence rate which increases with the learning rate, as expected. However, it is interesting to note that in this setting, a lower loss does not translate to better evaluation results. As can be seen in Figure~\ref{fig:result_lr} (left), the test score with the lowest Learning Rate (i.e., 0.0001) is close to 0, which is not the case of the other Learning Rate values which obtained a similar loss score in the training phase, and even the experiment with oscillation behavior presents a better prediction performance. This behavior indicates that the slow convergence with the lowest Learning Rate can make the model lose generalization and overfit the training data. 

\begin{figure}[htpb]
    \centering
    \includegraphics[width=0.85\textwidth,height=7cm]{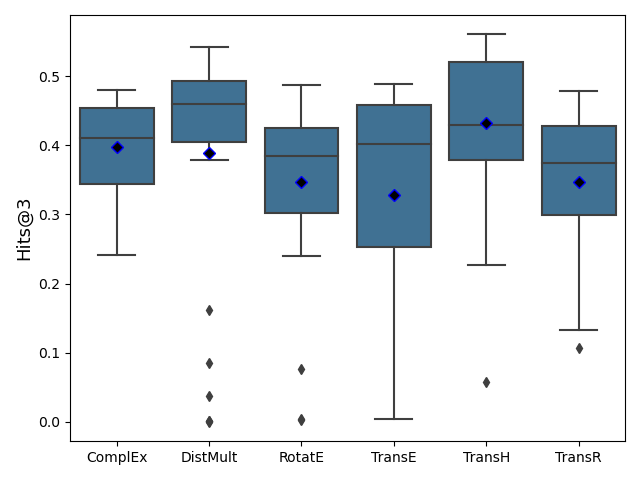}
  \caption{Boxplot summary of the Hits@3 score grouped by Model.}
  \label{fig:result_model}
\end{figure}

\vspace{\baselineskip}
Finally, we can compare the global evaluation results for the different models among all the experiments. Figure~\ref{fig:result_model} shows a  graphical summary of the Hits@3 performance. It can be seen that the best overall mean performance is obtained by TransH, which is also the model that obtained the highest evaluation score. It is followed by ComplEx and DistMult, which have a very similar mean value, and although the former is designed as an improvement to the latter, DistMult has a higher median. The worst performing model is TransE, the simplest one, which for multiple configurations is not able to capture the graph information. 

\vspace{\baselineskip}
With the results of the experiments, an automatic hyperparameter tuning optimization has been carried out to obtain embeddings with a better performance than those found in the previous experiments. The model chosen for this step has been the translational model TransH, as it has generated the best performing model and the highest mean performance. The hyperparameter optimization has been performed using the Tree-Structured Parzen Estimator (TSE) algorithm, which sequentially constructs probability models from which to draw the hyperparameter configuration based on past performances, aiming to explore the promising areas of the search space. As seen in Table~\ref{tab:hyperopt}, after 100 iterations of the hyperparameter tuning optimization for TransH, the resulting embeddings slightly increased the Hits@3 performance (2.03\%). Therefore, the best results have been obtained for hyperparameter configurations that are compliant with the results previously found. 

\begin{table*}[]
\centering
\caption{Hyperparameter optimization final configuration and results.}
\label{tab:hyperopt}
\begin{tabular}{l c}
\toprule
 & \textbf{TransH} \\
 \midrule
\textbf{Embedding Dimension} & 42 \\
\textbf{Learning Rate} & 0.0012 \\
\textbf{NPP} & 50 \\
\textbf{Hits@3} & 0.5727 \\
\bottomrule
\end{tabular}
\end{table*}

\subsubsection{Domain Expert Validation}
The LP results over the whole KG have been used to compare different models, to observe how their performance changes with different hyperparameters and to have a general idea of how well the KG has been captured. Additionally, although there is no data available for an offline evaluation of the recommendations, the suggestions produced by the LP anticipation system have been qualitatively assessed by a ML expert. For that, 3 classification and 3 regression datasets, which the system has not previously seen, have been fed to the LP engine, as if a new user was requesting workflows for them (see Table~\ref{tab:summary}). Then, all the entities have been scored for the different relations to be anticipated. The decision to not filter the entities (i.e., rank all the entities, not only those compatible with the relation) has been made to also assess the capability of the learned KGE to capture the graph structure. Finally, the top three scoring entities have been presented to the expert, who has assessed them in terms of sensibility and relevance according to the dataset for which they have been produced.

\subparagraph{Intent}
Regarding the Intent, the system anticipated as a first option in all cases the Intent related to the expected analysis, and the second and third recommendations were also of class \textit{Intent}. This recommendation could also have been done with an heuristic rule inspecting the nature of the target variable, similarly to what can be done with the query anticipation system. Therefore, given that this characteristic is annotated in the graph, it suggests that this relationship is correctly captured by the KGE. However, the expert pointed out that some users could want to, for instance, discretize a numeric target variable during the preprocessing step, which would result in giving an incorrect suggestion if it is only based on the target's variable type. Therefore, it is clear that this suggestion should not be done solely by taking into account the variable's type, hence by using KGEs this error can be mitigated as extra information can influence the recommendation.
\subparagraph{Evaluation Requirements}
Regarding the anticipation of Evaluation Requirements, once again all the top three recommended entities are relevant, as they correspond to instances of class \textit{EvaluationRequirements}. Additionally, the suggested entities match the Intent (e.g., F1-Score for classification datasets) for all the experiments with the exception of one of the three regression datasets, where the last suggestion is not relevant to the expected Intent. This behavior can be explained by the fact that the amount of classification datasets is twice the amount of regression datasets present in the experiments, and will not be given by the recommendation engine once the candidates are filtered. With respect to the relevance of the other suggestions, the expert pointed out that they are all sensible and valid, and depend on a more concrete evaluation goal (e.g., if the user’s purpose is to reduce the False Positives in a Classification task, the Precision metric may be prioritized).

\begin{sidewaystable}[htbp]
\caption{Recommendation output for the domain expert validation of the LP model. For each dataset, the top three intents, metrics and constraints are provided. The errors identified by the expert are highlighted in bold.}
\label{tab:summary}
\small
\begin{tabular}{cccccccc}
\toprule
Dataset & Variables & Instances & Expected Analysis & Intent & Metric & Constraints\\
\midrule
Seeds & 7 & 211 & Classification & Classification & F1-Score & SVC & \\
\addlinespace
& & & & Regression & Accuracy & KNeighborsClassifier & \\
\addlinespace
& & & & Clustering & AUC & LogisticRegression & \\
\midrule
Cancer & 29 & 569 & Classification & Classification & Precision & RandomForest & \\
\addlinespace
& & & & Regression & Accuracy & SVC & \\
\addlinespace
& & & & Clustering & F1-Score & \textbf{NoPreprocessing} & \\
\midrule
Iris & 4 & 150 & Classification & Classification & Accuracy & SVC & \\
\addlinespace
& & & & Regression & F1-Score & LogisticRegression & \\
\addlinespace
& & & & Clustering & Precision & RandomForest & \\
\midrule
Boston & 14 & 506 & Regression & Regression & R2 & SVR & \\
\addlinespace
& & & & Classification & MSE & SGDRegressor & \\
\addlinespace
& & & & Clustering & RMSE & KNeighborsRegressor & \\
\midrule
MPG & 7 & 398 & Regression & Regression & RMSE & SVR & \\
\addlinespace
& & & & Classification & R2 & Normalizer & \\
\addlinespace
& & & & Clustering & MSE & MLPRegressor & \\
\midrule
Car Price & 15 & 159 & Regression & Regression & RMSE & SVR & \\
\addlinespace
& & & & Classification & RMSE & RandomForestRegressor & \\
\addlinespace
& & & & Clustering & \textbf{F1-Score} & KNeighborsRegressor & \\
\bottomrule
\end{tabular}
\end{sidewaystable}

\subparagraph{Constraints}
With regard to the anticipation of Constraints, the behavior is similar to the one of Evaluation Requirements. First of all, all the recommended entities are members of the appropriate class (i.e., \textit{Constraint}) and the  recommendations adjust to the Intent’s expected purpose. The recommended constraints are sensible, as they recommend well known and powerful algorithms, but depend again on the user’s more fine grained goal. However, the expert pointed out that for one dataset the constraint \textit{NoPreprocessing} was suggested, which is usually not advised to constrain unless there is a strict requirement for it. The reason for this recommendation is that during the automatic creation of workflows to populate the KG, the option of not using a preprocessing step appears frequently, as the synthetic user selected the constraints without any restriction.  

\vspace{\baselineskip}

\noindent Overall, the experiments show that the learned KGEs are a promising solution for capturing the KG, and that by using the LP method, relevant suggestions are able to be presented to the user.

\section{Prototype}
\label{sec:prototype}

\begin{figure*}[htpb]
    \centering
    \includegraphics[width=0.85\textwidth]{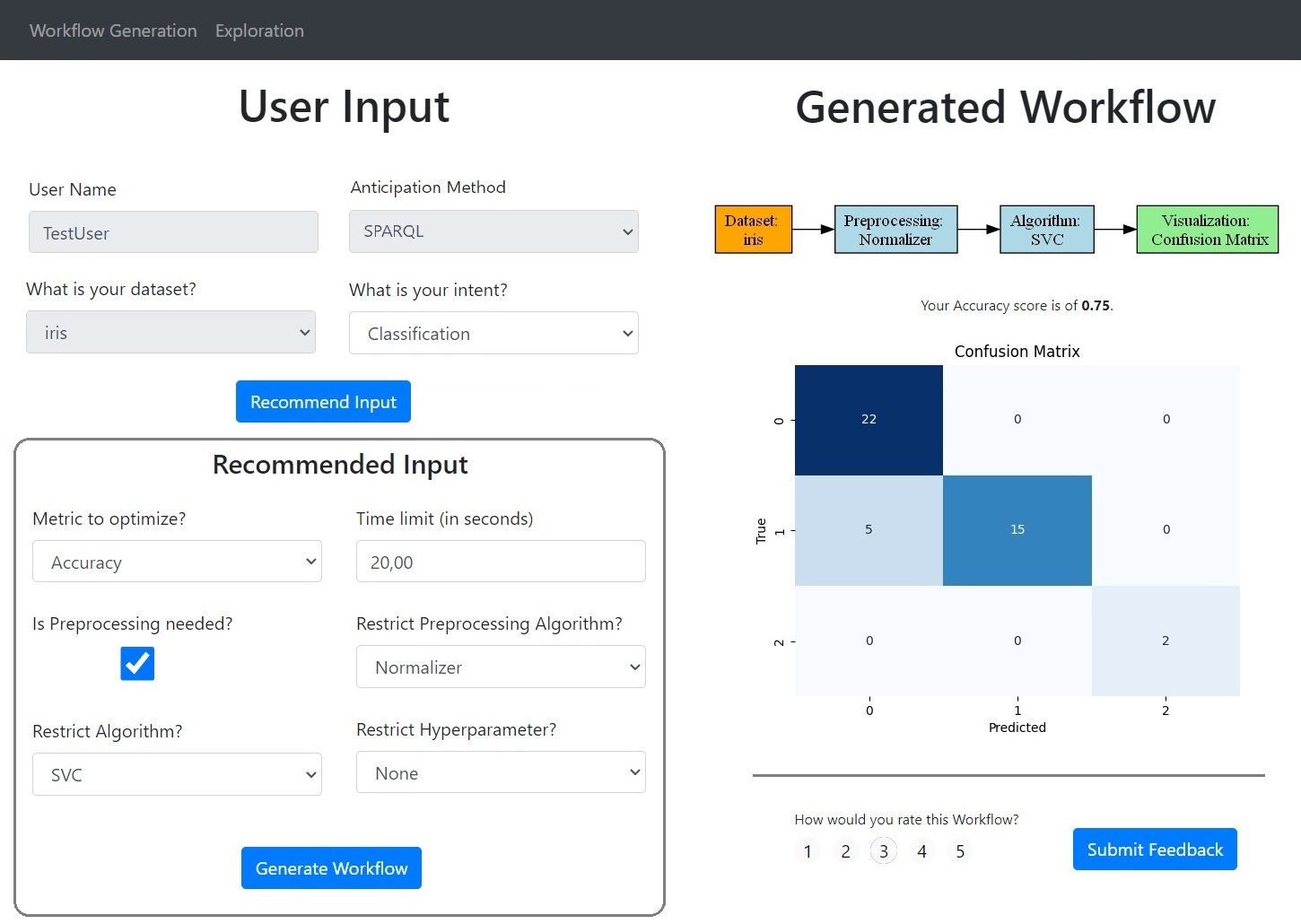}
  \caption[User Input]{Example of the assisted user input (left) and the generated workflow with the results (right) in the prototype.}
  \label{fig:UI}
\end{figure*}

Finally, we developed a prototype\footnote{https://github.com/gerardponsrecasens/capturing-and-anticipating-ML-intents} to show the feasibility of our approach, which has been created as a user interface for an AutoML tool~\cite{hyperopt}. Our system assists users in specifying the ML workflows they would like to execute. The proposed interface for this tool is shown in Figure~\ref{fig:UI}. As can be seen, users are assisted while stating their intent, the evaluation requirements and their constraints, by getting ordered suggestions resulting from the two approaches presented in this work.

The interaction is as follows: First of all, the users select the dataset they want to use (see upper-left part of Figure~\ref{fig:UI}). If the dataset has not been previously used by the system, its relevant characteristics are extracted and annotated. Next, the users have to choose their Intent. Depending on the anticipation method selected, predefined query templates will be run to retrieve the most appropriate Intent (as explained in Section~\ref{sec:queries}) or a LP task will be executed by ranking the \textit{(Task, hasIntent, ?)} triple (see Section~\ref{sec:embeddings}). After selecting the Intent, the users are recommended the evaluation requirements and algorithm constraints (see bottom-left part of Figure~\ref{fig:UI}). Again, either the result of SPARQL queries or LP recommendations will be provided to the user. Once the users submit all the inputs, the workflow generator creates an optimal workflow with the allocated time and taking into account the constraints, as explained in Section~\ref{sec:workflow_instantiation}. The workflow generator used is HyperOpt~\cite{hyperopt}, hence the users’ inputs are mapped into the appropriate Python scripts that allow the execution of the workflow generator. If one wishes to swap the AutoML module, the mapping needs to be adjusted accordingly. The users are then presented with the Workflow results, in terms of the score of the chosen metric, a related visualization and a high level view of the generated workflow (see right part of Figure~\ref{fig:UI}). Feedback can also be provided in this step, as can be seen in the bottom-right part of Figure~\ref{fig:UI}. After the feedback is submitted, the whole interaction of the user is annotated (i.e., dataset, intent, constraints, requirements and feedback) and the generated workflow is parsed to the appropriate triples, which are then incorporated to the KG. 

For the query based anticipation approach, incorporating the triples to the KG after each interaction is enough for the system to provide up to date recommendations. However, for the LP task the embeddings should be fine-tuned after every usage of the system so that the recently generated workflows are taken into account. Given the small set of new facts that need to be incorporated in the KG for every interaction, this tuning step does not create a significant overhead on the system's functionality. Additionally, as usual in RSs, the model (in our case, the embeddings) should be periodically retrained to better capture up-to-date results. This retraining is influenced by the size of the KG which increases with each interaction with the system. Therefore, in the long run, different strategies could be designed to filter the data with which the system is trained in terms of quality (e.g., expertise of the user, results, etc.) or relevance (e.g., usage of state-of-the-art models). This effect is not present during inference time, as the candidates are filtered based on the range of the relation, hence the candidate pool only grows when new algorithms or processes are added to the system.

\section{Conclusions and Future Work}
\label{sec:conclusion}
The first objective of this work was to create a KG capable of capturing the entirety of data analytics processes, from the different workflow steps and their characteristics to the user’s inputs and feedback. To this end, new concepts such as Constraints, Preferences, and Intents have been introduced to extend and connect existing ontologies (DMOP and Person). 

The resulting KG has been populated in accordance with its schema from a variety of sources, and has been used to explore the second goal, which is the anticipation of user intentions. For that, the KG has been first leveraged for knowledge extraction through queries by creating query templates capable of retrieving the necessary information to provide recommendations. To solve the limitations posed by the querying method, the usage of knowledge graph embeddings for recommendation tasks with link prediction has been implemented, assessing its performance by the ability to capture the whole KG. Different experiments have shown that this approach can successfully capture the graph structure and give sensible recommendations through LP.

Regarding future work, the exploitation of the KG by different modules of DA Assistants could be explored. For instance, the KG could interact with the workflow generator and modify the exploration of the search space according to the stored data (see Figure~\ref{fig:general}, arrow 5). Additionally, the KG could be expanded to model processes such as data integration or security protocols, or to include more information about the input data. Finally, an extra layer of abstraction could be designed for the intent hierarchy, aiming to translate business oriented terms and questions to the existent intent terminology. 



\bibliographystyle{elsarticle-num} 
\bibliography{references}






\end{document}